\documentclass[runningheads]{llncs}
\usepackage[T1,T2A]{fontenc}
\usepackage[utf8]{inputenc}
\usepackage[english,russian]{babel}
\usepackage{multirow}
\usepackage{float}
\usepackage{hyperref}
\usepackage{array}
\usepackage{graphicx}

\begin{document}
\title{Evaluation of Sentence Embedding Models for Natural Language Understanding Problems in Russian}
\titlerunning{Sentence Embeddings for Russian NLU}
%
\author{Dmitry Popov \and Alexander Pugachev \and Polina Svyatokum \and Elizaveta Svitanko \and Ekaterina Artemova}
\authorrunning{D. Popov et al.}
%
\institute{National Research University Higher School of Economics\\
\email{\{dgpopov, avpugachev, posvyatokum, eisvitanko\}@edu.hse.ru}, \email{echernyak@hse.ru}}

\maketitle              
\selectlanguage{english}
\begin{abstract}
We investigate the performance of sentence embeddings models on several tasks for the Russian language. In our comparison, we include such tasks as multiple choice question answering, next sentence prediction, and paraphrase identification. We employ FastText embeddings as a baseline and compare it to ELMo and BERT embeddings. We conduct two series of experiments, using both unsupervised (i.e., based on similarity measure only) and supervised approaches for the tasks. Finally, we present datasets for multiple choice question answering and next sentence prediction in Russian. 

\end{abstract}

\section{Introduction} 

With word embeddings have been in the focus of researchers for several decades, sentence embeddings recently started to gain more and more attention. A word (sentence) embedding is a projection in a vector space of relatively small dimensionality, that can capture word (sentence) meaning by making the embeddings of two words (sentences) that are similar to get closer in this vector space. With no doubts, the usage of the properly trained word embeddings boosted the quality of the majority of natural language processing (NLP), information extraction (IE), neural machine translation (NMT) tasks. However, it seems that word embedding models are facing their limits when it comes to polysemy and ambiguity. A natural solution to these problems lies in sentence embedding models too, as they allow to capture the context--dependent meaning of any part of the sentence. 

For the last two years, the amount of projects and papers on sentence embeddings has increased dramatically. Several research groups show that a complex pre--trained language model may serve both as an input to another architecture and as a standalone sentence embedding model. The most famous models of this type, ELMo, and BERT, named after characters of Sesame Street show, can be treated as ``black boxes'', that read a sentence in and output a vector representation of the sentence. The efficiency of these models for the English language is well studied already not only in several natural language understanding (NLU) tasks but also for language modeling and machine translation.  However little has been done to explore the quality of sentence embeddings models for other languages, including Russian, probably due to the absence of NLU datasets.
The contribution of this paper is two--fold. \textbf{First}, we create two novel NLU datasets for the Russian language: a) multiple choice question answering dataset, which consists of open domain questions on various topics; b) next sentence prediction dataset, that can be treated as a kind of multiple choice question answering. Given a sentence, one needs to choose between four possible next sentences. Correct answers are present in both datasets by design, which makes supervised training possible. \textbf{Second}, we evaluate the quality of several sentence embedding models for three NLU tasks: multiple choice question answering, next sentence prediction and paraphrase identification. 

The results confirm, that in the tasks under consideration, sentence--level representations perform better than the word--level ones as in many other tasks.  

The remainder is structured as follows. Section 2 presents an overview of related works; Section 3 introduces the datasets, Section 4 describes the methods we used to tackle the tasks. The results of the experiments are presented in Section 5. Section 6 concludes.

\section{Related work}

A word embedding is a dense vector representation of a word that allows modeling some sort of semantic (or functional) word similarity. Two words are considered similar if a similarity measure, such as cosine function, for example, between corresponding vectors is high enough. As this definition is rather vague, there are two main approaches to evaluating the quality of a word embedding model. Intrinsic evaluation is based on conducting on standard word pairs and analogies datasets, such as Word--353\footnote{\url{http://www.cs.technion.ac.il/~gabr/resources/data/wordsim353/}} or Simlex--999\footnote{\url{https://fh295.github.io/simlex.html}}. External evaluation requires an external machine learning tasks, such as sentiment classification or news clustering, which can be evaluated by a quality measure, such as accuracy or Rank index. All factors of machine learning algorithms are held equals so that these quality measures are affected by the word embeddings model only.

The methodology of creation and evaluation of sentence embeddings is less developed, when compared to the zoo of word embedding models.  The evaluation of sentence embeddings models is usually conducted of natural language understanding tasks, such as semantic textual similarity, natural language inference, question answering, etc. Further, we overview the basic and more advanced models of sentence embeddings.

\subsection{Unsupervised sentence embeddings}

The simplest way of obtaining the sentence embedding is by taking the average of the word embeddings in the sentence. The averaging can treat words equally, rely on $tf-idf$ weights \cite{arroyo2019unsupervised},  take the power mean of concatenated word embeddings \cite{ruckle2018concatenated} and employ other weighting and averaging techniques. 

Another approach of unsupervised training of sentence embeddings is Doc2Vec  \cite{doc2vec}, which succeeds after Word2Vec and FastText.  Word2Vec \cite{w2v} by Mikolov et al. is a predictive embedding model and has two main neural network architectures: continuous Bag--of--Words (CBoW) and continuous skip--gram \cite{skip-gram}. Given a central word and its context (i.e., $k$ words to the left and $k$ words to the right), CBoW tries to predict the context words based on the central one, while skip--gram on tries to predict the context words based on the central one. 

Joulin et al. \cite{fasttext} suggest an approach called FastText, which is built on Word2Vec by learning embeddings for each subword (i.e., character $n$--gram, where $n$ can vary between some bounds and is a hyperparameter to the model). To achieve the desired word embeddings, the subword embeddings are averaged into one vector at each training step. While this adds a lot of additional computation to training, nevertheless it enables word embeddings to encode sub--word information, which appears to be crucial for morphologically rich languages, the Russian language being one of them. 

The goal of the aforementioned Doc2Vec approach, introduced by Mikolov et al. \cite{doc2vec}, is to create an embedding of a document, regardless of its length.

All the studies as mentioned earlier work properly on the English language and some of them release pre--trained Russian embeddings, too. Russian--specific RusVectores \cite{rusvectores} pre--trained model, which was trained on Russian National Corpora, Russian Wikipedia and other Web corpora possesses several pre--trained word embeddings models and allows to conduct a meaningful comparison between the models and their hyperparameters.  

The Skip--Thought model follows the skip--gram approach: given a central sentence, the context (i.e., the previous and the next) sentences are predicted. The architecture of Skip--Thought consists of a single encoder, which encodes the central sentence and two decoders, that decode the context sentences. All three parts are based on recurrent neural networks. Thus their training is rather difficult and time--consuming. 
\subsection{Supervised sentence embeddings}
In recent years, several sources say the unsupervised efforts to obtain embeddings for larger chunks of text, such as sentences, are not as successful as the supervised methods. Conneau et al. \cite{conneau} introduced the universal sentence representation method, which works better than, for instance, Skip--Thought  \cite{kiros2015skipthought} on a wide range of transfer tasks. It's model architecture with BiLSTM network and max--pooling is the best current universal sentence encoding method. The paper on Universal Sentence Encoder \cite{cer2018universal} discovers that transfer learning using sentence embeddings, which tends to outperform the word level transfer. As an advantage, it needs a small amount of data to be trained in a supervised fashion.  

\subsection{Language models}
One of the recently introduced and efficient methods are embeddings from Language Models (ELMo) \cite{elmo} that models both complex characteristics of word use, and how it is different across various linguistic contexts and can also be applied to the whole sentence instead of the words. Bidirectional Encoder Representations from Transformers (BERT) \cite{bert} has recently presented state--of--the--art results in a wide variety of NLP tasks including Natural Language Inference, Question Answering, and others. The application of an attention model called Transformer allows a language model to have a better understanding of the language context. In comparison to the single--direction language models, this one uses another technique namely Masked LM (MLM) for a bidirectional training. 

\subsection{Evaluation of sentence embedding models}

While different embedding methods are previously discussed, the most suitable evaluation metric is also a challenge. Quality metrics are largely overviewed in RepEval\footnote{\url{https://aclweb.org/anthology/events/repeval-2017/}} proceedings. In \cite{bakarov2018survey}  both widely--used and experimental methods are described. SentEval is used to measure the quality of  sentence representations for the tasks of  natural language inference or sentence similarity \cite{conneau2018senteval}.
The General Language Understanding Evaluation (GLUE) benchmark\footnote{\url{https://gluebenchmark.com}} is one of the popular benchmarks for evaluation of natural understanding systems. The top solutions, according to the GLUE leadership, exploit some sort of sentence embedding frameworks. GLUE allows testing any model in nine sentence or sentence--pair tasks, such as natural language inference (NLI), semantic textual similarity (paraphrase identification, STS) or question answering (QA).

\section{Datasets}

\subsection{Multiple Choice Question Answering (MCQA)}
The portal \href{https://geetest.ru/}{geetest.ru} provides tests on many subjects such as history, biology, economics, math, etc., with most of the tasks being simple wh--questions. These tests were downloaded to create a multiple choice question answering dataset.

Every test is a set of questions in a specific area of a certain subject.
For the final dataset we handpicked tests from the following subjects: Medicine, Biology, History, Geography, Economics, Pedagogy, Informatics, Social Studies.

The selection was based on two criteria. Firstly, questions should be answerable without knowing the topic of the test. For example, some questions in test could not be answered correctly without presenting a context of a specific legal system. Secondly, questions should test factual knowledge and not skills. For example, almost any math test will require to perform computations, and such type of task is not suitable for this dataset.

What is more, we have collected questions on history and geography from \href{https://ege.sdamgia.ru}{ege.sdamgia.ru}. The selection of questions was similar to described above.
As a result, the total number of questions is around 11k with three subjects being larger than other. These subjects are: medicine, (4k of questions), history (3k of questions), biology (2k of questions). The resulting dataset is somewhat similar to  Trivia QA dataset, however the domains are different \cite{triviaqa}. 

\subsection{Multiple choice next sentence prediction (NSP)}
We have collected a new dataset with 54k multiple choice questions where the objective is to predict the correct continuation for a given context sentence from four possible answer choices. The dataset was produced using the corpora of news articles of ``Lenta.ru''\footnote{\url{https://github.com/yutkin/Lenta.Ru-News-Dataset}}. To sample correct and incorrect answer choices we chose a trigram and a context sentence which ends on this trigram. The correct answer choice was the continuation of the context sentence, and the incorrect choices were the sentences following the trigram in other sentences in the corpora. Labels of correct answers were sampled uniformly. So, the random or constant predictions results in an accuracy score of 0.25.

\subsection{Paraphrase identification (PI)}
For a paraphrase detection task, we used Russian language paraphrase dataset collected from news titles \footnote{\url{http://paraphraser.ru}}. The dataset contains 7k pairs of titles which are the same, close and different by meaning. Constant prediction on this dataset gives us an accuracy score of 0.64. In a sense, Microsoft Research Paraphrase Corpora\cite{microsoft_paraphrase} is similar to this dataset. Both of them were collected from news titles.

\begin{table}[h]
    \small
    \selectlanguage{russian}
    \centering
    \begin{tabular}{p{12cm}}
    \hline
    \textbf{Question} \newline
    Какие из указанных симптомов характерны для фарингита?  \\ [0.5ex]
    \textbf{Answer choices} \newline
    \textbf{1. резкая боль в горле} \\
    2. першение и дискомфорт в горле \\
    3. затруднение проглатывания слюны \\
    4. субфебрильная температура \\ [0.5ex]
    \hline
    \end{tabular}
    \selectlanguage{english}
    \caption{Examples from MCQA dataset. The correct answer is bolded.}
    \label{tab:mcqa_ex}

    \small 
    \selectlanguage{russian}
    \centering
    \begin{tabular}{p{12cm}}
    \hline
    \textbf{Context} \newline
    Мартин Скорсезе намеревается приступить к съемкам экранизации романа «Молчание» японского писателя Сюсаку Эндо в 2014 году,  \\ [0.5ex] 
    \textbf{Answer choices} \newline
    \ 1. был оснащен одиннадцатиметровым стеклянным полом шириной два метра, сообщается в полученном «Домом» пресс-релизе компании «Сен-Гобен». \newline
    \ 2. когда он провозил мак для одной из продуктовых баз \newline
    \ \textbf{3. сообщает Deadline.  Финансированием проекта займутся компании Emmett/Furla Films и Corsan Films} \newline
    \ 4. а производить трубы там начали уже спустя два года.  В числе поставщиков «Газпрома» ЗТЗ появился в 2017 году \\ [0.5ex]
    \hline
    \end{tabular}
    \selectlanguage{english}
    \caption{An example from NSP dataset. The correct answer is bolded.}
    \label{tab:nsp_ex}
%
%
%
%
%
    \centering
    \selectlanguage{russian}
    \begin{tabular}{|p{5.5cm}|p{5.5cm}|c|}
    \hline
         text 1 & text 2 & label  \\
         \hline
         Мэрилин Мэнсон передумал выступать в России. & Мэрилин Мэнсон отменил тур по России. & 1 \\ \hline
         Бывший чемпион мира по боксу умер в 48 лет. & Как судей судили за их решения. & 0 \\
         \hline
        
    \end{tabular}
    \selectlanguage{english}
    \caption{Examples from the PI dataset.}
    \label{tab:my_label}
\end{table}



\subsection{Dataset statistics}

\begin{figure}
  \caption{Frequency distribution of top 25 tokens for MCQA, NSP, PI datasets.}
    \label{fig:data_top_25}
  \centering
    \includegraphics[width=0.3\textwidth]{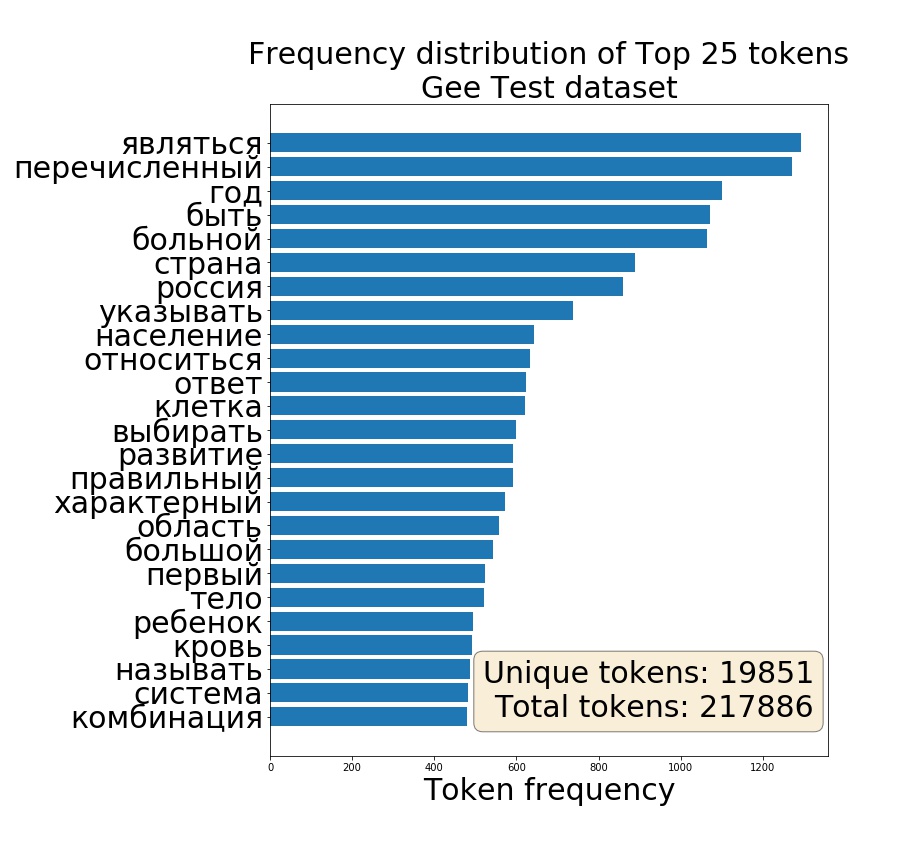}
    \includegraphics[width=0.3\textwidth]{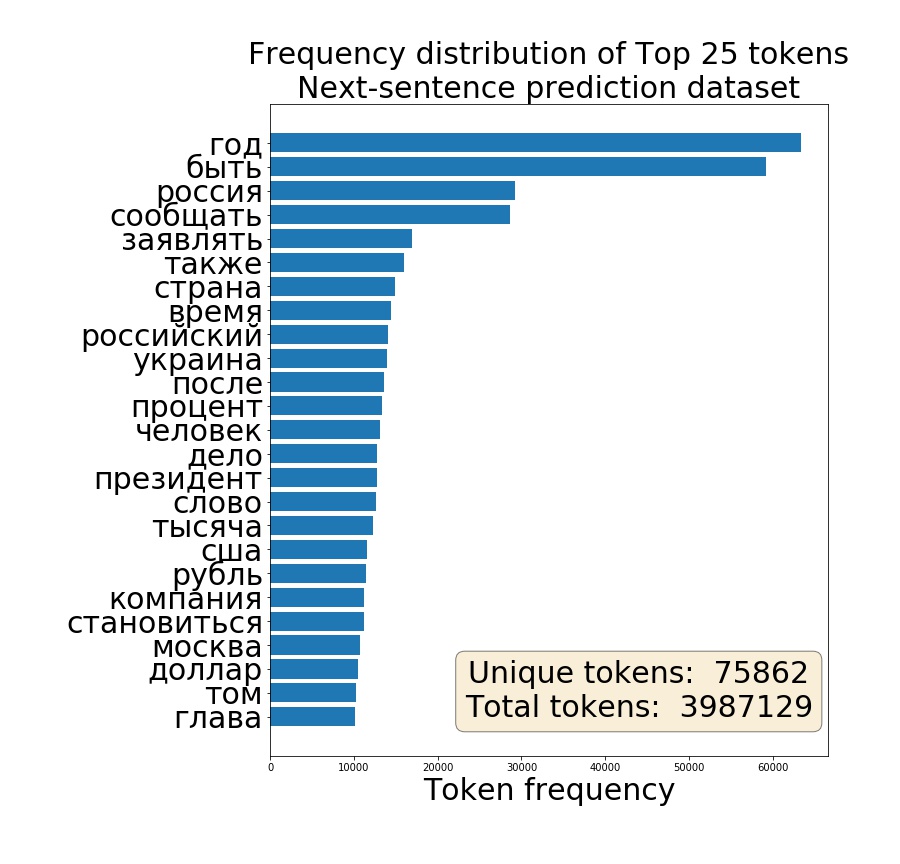}
    \includegraphics[width=0.3\textwidth]{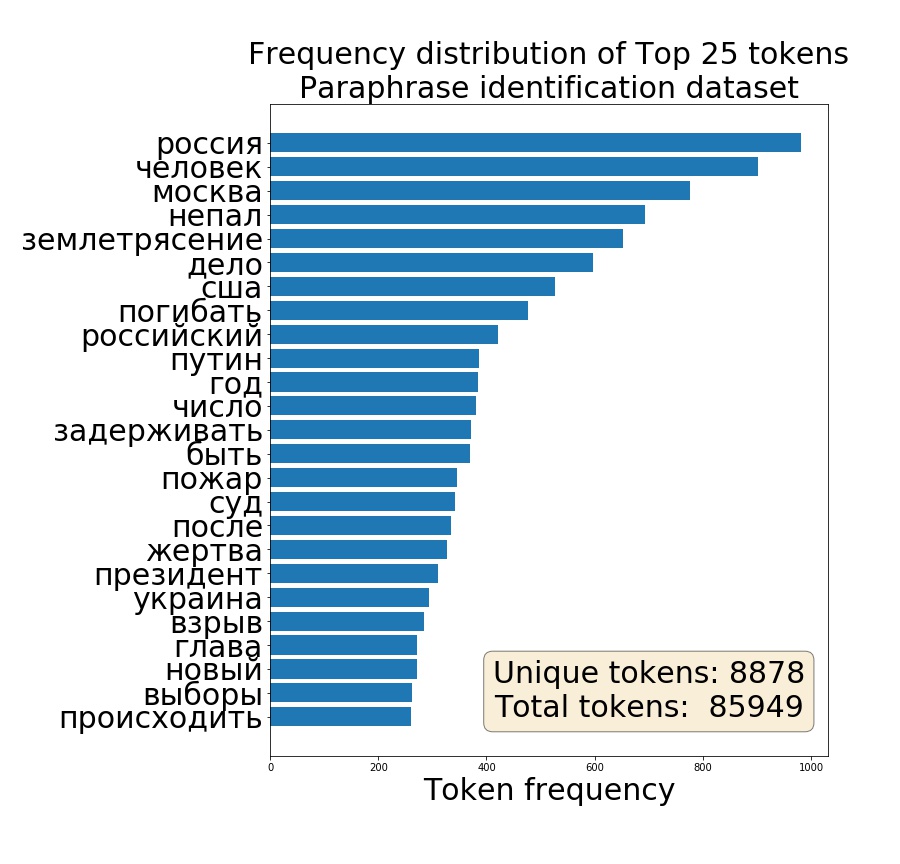}
\end{figure}

\begin{figure}
  \caption{Sentence length distribution for MCQA, NSP, PI datasets.}
    \label{fig:data_len_hist}
  \centering
    \includegraphics[width=0.3\textwidth]{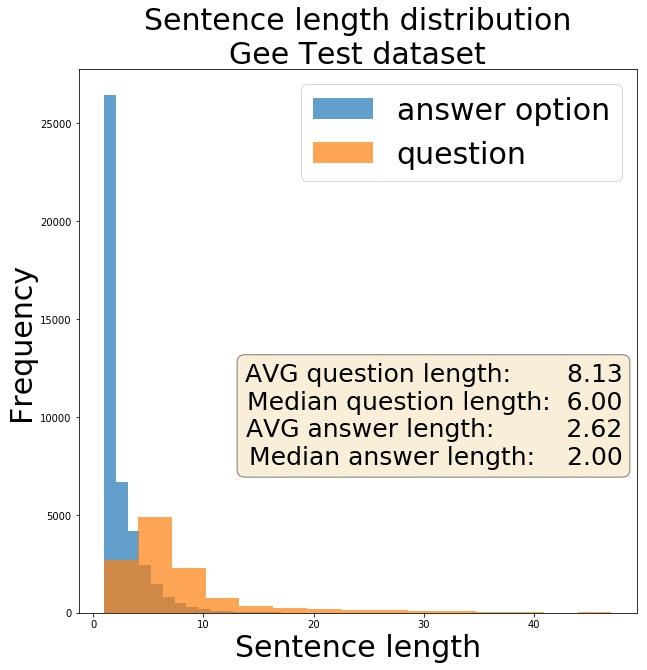}
    \includegraphics[width=0.3\textwidth]{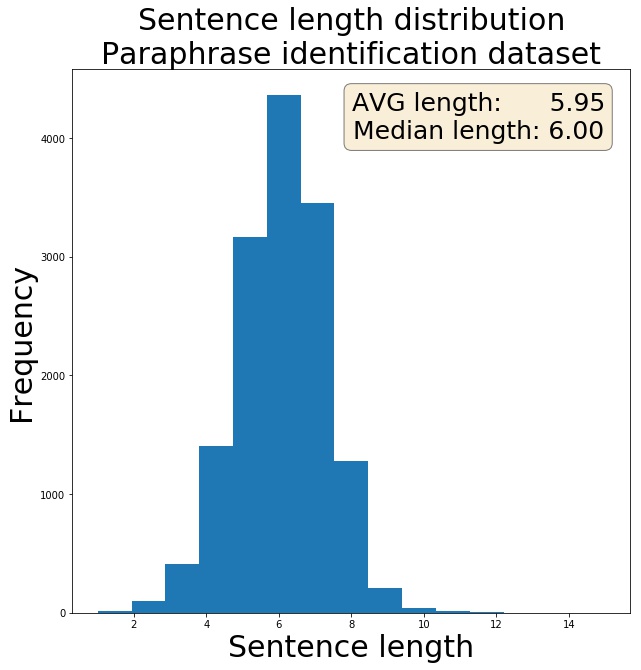}
    \includegraphics[width=0.3\textwidth]{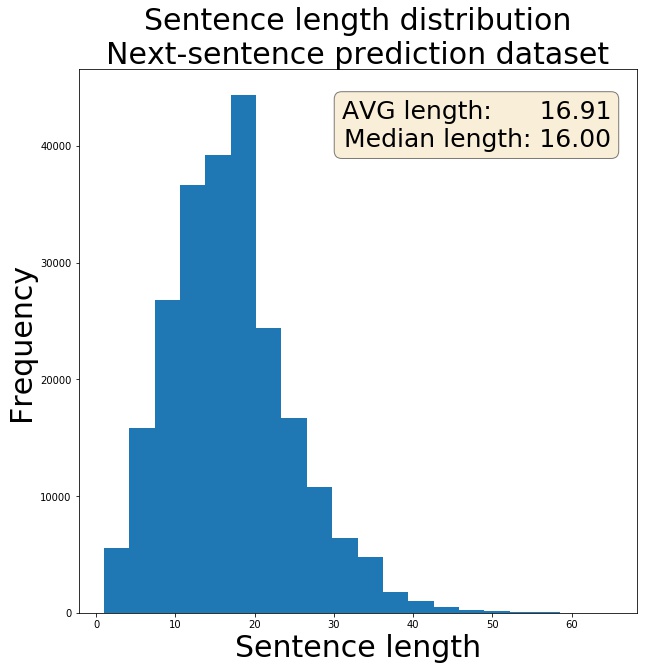}
\end{figure}

Frequency distribution of top 25 most frequent tokens, the number of unique tokens and the total number of tokens in the datasets can be found in Figure \ref{fig:data_top_25}. Sentence length distribution, average and median sentence length can be found in Figure \ref{fig:data_len_hist}.

\section{Methods} 

There are two types of problems we were considered for the comparison of sentence embeddings:
\begin{itemize}
    \item \textbf{Multiple choice questions.} Datasets for this type are MCQA and NSP ones. The objective of the problem is to predict the correct answer for a given question/context and four answer choices.
    \item \textbf{Paraphrase identification.} The goal for such a problem is to predict if two given sentences are a paraphrase or not.
\end{itemize}

To compare different methods of obtaining sentence embeddings, we have explored supervised and unsupervised scenarios of using such embeddings.
\subsection{Unsupervised approach}
In unsupervised methods, we are interested in a similarity between sentence embeddings in terms of cosine similarity.

\textbf{Multiple choice questions.}
For such type of problem for a given set $\{q, a_1,$  $a_2, a_3, a_4\}$, where $q$ is an embedding of either a question or a context sentence and $a_i$ is an embedding of $i$--$th$ answer choice, the predicted answer choice is the choice which embedding is the most similar to $q$.

\textbf{Paraphrase identification.} Let for a given pair of sentences $t_1$ and $t_2$ are sentence embeddings of this pair respectively. First of all, we split the dataset into training and test sets, after that searching for a threshold $t$ on a training set such that pairs with $sim(t_1, t_2) > t$ will be labeled as a paraphrase. Finally, we will evaluate results on a test set. 

\subsection{Supervised approach}
Text vector representations are often inputs to machine learning models. In this approach, we are aiming to figure out which methods of obtaining vector representations are better as inputs into linear models such as the logistic regression and which methods are better as inputs to a gradient boosting models such as the CatBoost \cite{catboost}.

\textbf{Multiple choice questions.} Since we cannot just run a multiclass classification as the correct answers numbers are not related to questions we will make a binary classification model which predicts a probability for a given question--answer (context--answer) pair to be correct, i.e. the answer is the correct answer choice for the question/context.  Then for a given question/context and four answers, the predicted answer is the answer such that the model gives the highest probability.

\textbf{Paraphrase identification.} For this problem, we just build a binary classifier on a concatenation of sentence embeddings.

\section{Experiments and results} 
\subsection{FastText}

\begin{table}[]
\centering
\selectlanguage{english}
\caption{FastText embeddings results.}
\label{tab:fasttext_results}
\begin{tabular}{ c  c  c  c  c  c  c }

\hline
   \multirow{2}{*}{Method} & \multicolumn{2}{c}{\textbf{MCQA}} & \multicolumn{2}{ c }{\textbf{NSP}} &  \multicolumn{2}{c}{\textbf{PI}}\\ 
 & \ Accuracy \ & \ \ F1 \  \  &  \ \  Accuracy \ \  & F1 & \ \  Accuracy \ \  & F1 \\
\hline
Unsupervised      & 0.305    & 0.304 & 0.337 & 0.337 & 0.719     & 0.806  \\
Logistic regression           & 0.287    & 0.287  & 0.248 & 0.248 & 0.704 & 0.612 \\
CatBoost     & 0.318    & 0.317  & 0.496 & 0.496 & 0.762 & 0.715  \\
\hline
\end{tabular}

\end{table}

As it can be seen from Table \ref{tab:fasttext_results}, FastText model accomplishes differently depending on tasks and methods. The best quality on each task was reached by CatBoost method according to Accuracy metrics. According to the F1 score, CatBoost also outperforms other methods in all of the tasks except Paraphrase identification where the best score with a large margin was achieved by the unsupervised method. There can also be seen a big quality difference between CatBoost and other methods in next sentence prediction task. Logistic regression shows the worst results in each task. The overall performance of methods based on FastText model is far from perfect. The results achieved using FastText can be considered as a baseline for our investigation.

\subsection{ELMo}
We have used three ELMo models \footnote{\url{http://docs.deeppavlov.ai/en/master/intro/pretrained_vectors.html}} pre--trained on Russian Wikipedia, ``Lenta.Ru'' news articles and Russian tweets corpora, respectively. So, one of the main results obtained from our experiments is how different domains of pre--trained models affects final results.
\begin{table}[htp!]
\centering
\selectlanguage{english}
\caption{ELMo unsupervised experiments results.}
\begin{tabular}{m{2cm} c c c c c c c}
\hline
   \ \ \ \multirow{2}{*}{Method} & \multirow{2}{*}{Domain} & \multicolumn{2}{c}{\textbf{MCQA}} & \multicolumn{2}{ c }{\textbf{NSP}} &  \multicolumn{2}{c}{\textbf{PI}}\\ 
 & & \ Accuracy \ & \ \ F1 \  \  &  \ \  Accuracy \ \  & F1 & \ \  Accuracy \ \  & F1 \\
\hline
\multirow{3}{*}{Unsupervised} & Wikipedia      & 0.300    & 0.300 & 0.645 & 0.645 & 0.807 & 0.867  \\
& News           & 0.293    & 0.293  & 0.691 & 0.691 & 0.807 & 0.866    \\
& Twitter     & 0.291    & 0.291  & 0.559 & 0.559 & 0.803 & 0.863  \\
\hline
\multirow{3}{\hsize}{Logistic regression} & Wikipedia      & 0.301    & 0.300 & 0.249 & 0.249 & 0.684 & 0.652  \\
 & News           & 0.318    & 0.318  & 0.249 & 0.248 & 0.702 & 0.668    \\
 & Twitter     & 0.317    & 0.316  & 0.251 & 0.250 & 0.705 & 0.674  \\
 \hline
\multirow{3}{\hsize}{CatBoost} &Wikipedia   & 0.314    & 0.314 & 0.647 & 0.647 & 0.773 & 0.729  \\
& News           & 0.310    & 0.310  & 0.669 & 0.669 & 0.797 & 0.758    \\
& Twitter     & 0.314    & 0.314  & 0.631 & 0.631 & 0.779 & 0.741 \\
\hline
\end{tabular}
\label{tab:elmo}
\end{table}

The results achieved by three methods based on three different pre--trained ELMo models are presented in Table \ref{tab:elmo}. The performance of these methods is quite sensitive to the source of training data for ELMo model.  For example, regarding the MCQA task, models trained on the News and Twitter corpora perform better than the model trained on Wikipedia, especially when logistic regression is used. In the unsupervised setting, the quality of next sentence prediction task highly depends on the source of training data ELMo model, too. However, in most cases there is no significant difference in results between three ELMo models. 

One can notice that logistic regression in both cases (FastText model and all three ELMo models) shows the worst results in the majority of the tasks. Regarding the next sentence prediction task, the performance of logistic regression has not gone far away from random choice. However, it shows better results than other models in the MCQA task. The unsupervised method achieved the best results for the paraphrase identification task. We can claim that the use of ELMo model contributed to better results in next sentence and paraphrase identification tasks, as there was observed significant improvement in accuracy and F1 scores. Speaking of MCQA task, the results are comparable with the previously obtained.

\subsection{BERT}

\begin{table}[htp!]
\centering
\selectlanguage{english}
\caption{BERT embeddings results on MCQA dataset.}
\begin{tabular}{ c  c  c m{2cm} c  c  c  c}

\hline
   \multirow{2}{*}{Method} & \multicolumn{3}{c}{\textbf{Best score}} & \multicolumn{2}{c}{\textbf{Average score}} &  \multicolumn{2}{c}{\textbf{Worst score}}\\ 
 & \ \ \ Accuracy \ & \ \ \ \ F1 \  \  &  \ \ \ \ \ Layer &\ \  Accuracy \ \  & F1 & \ \  Accuracy \ \  & F1 \\
\hline
Unsupervised      & 0.303    & 0.302 & Concatenation of layers from 1 to 6 & 0.292 & 0.292 & 0.274 & 0.274  \\ \hline
Logistic regression           & 0.336    & 0.335 & Layer number 1  & 0.324 & 0.323 & 0.310 & 0.310    \\ \hline
CatBoost     & 0.346    & 0.346  & Max pooling of layers from 4 to 6 & 0.326 & 0.326 & 0.312 & 0.311 \\
\hline
\end{tabular}
\label{tab:bert_mcqa}

\begin{center}
\selectlanguage{english}
\caption{BERT embeddings results on NSP dataset.}
\begin{tabular}{ c  c  c m{2cm} c  c  c  c }

\hline
   \multirow{2}{*}{Method} & \multicolumn{3}{c}{\textbf{Best score}} & \multicolumn{2}{ c }{\textbf{Average score}} &  \multicolumn{2}{c}{\textbf{Worst score}}\\ 
 &  \ \ \ Accuracy \ & \ \ \ \ F1 \  \  &  \ \ \ \ \ Layer &  \ \  Accuracy \ \  & F1 & \ \  Accuracy \ \  & F1 \\
\hline
Unsupervised      & 0.508    & 0.508  & Layer number 12 & 0.457 & 0.457 & 0.429 & 0.429  \\ \hline
Logistic regression           & 0.255    & 0.255  & Max pooling of layers from 7 to 9 & 0.249 & 0.248 & 0.244 & 0.244    \\ \hline
CatBoost     & 0.514    & 0.514  & Average pooling of layers from 7 to 12 & 0.479 & 0.479 & 0.414 & 0.414 \\
\hline
\end{tabular}
\label{tab:bert_nsp}
\end{center}

\begin{center}
\selectlanguage{english}
\caption{BERT embeddings results on PI dataset.}
\begin{tabular}{ c  c  c m{2cm} c  c  c  c }

\hline
   \multirow{2}{*}{Method} & \multicolumn{3}{c}{\textbf{Best score}} & \multicolumn{2}{ c }{\textbf{Average score}} &  \multicolumn{2}{c}{\textbf{Worst score}}\\ 
 &\ \ \ Accuracy \ & \ \ \ \ F1 \  \  &  \ \ \ \ \ Layer &  \ \  Accuracy \ \  & F1 & \ \  Accuracy \ \  & F1 \\
\hline
Unsupervised      & 0.801    & 0.857 & Average pooling of layers from 1 to 3 & 0.787 & 0.851 & 0.775 & 0.843  \\ \hline
Logistic regression           & 0.715  & 0.676 & Average pooling of layers from 7 to 12 & 0.693 & 0.651 & 0.670 & 0.615   \\ \hline
CatBoost     & 0.778    & 0.732 & Layer number 3  & 0.763 & 0.713 & 0.749 & 0.694 \\
\hline
\end{tabular}
\label{tab:bert_para}
\end{center}
\end{table}

\begin{table}[htp!]
\centering
\selectlanguage{english}
\caption{BERT embeddings results on MCQA dataset (1st and 12th layer).}
\begin{tabular}{ c  c  c c c  c  c  c}

\hline
   \multirow{2}{*}{Method} & \multicolumn{2}{c}{\textbf{1st layer}} & \multicolumn{2}{c}{\textbf{12th layer}} &  \multicolumn{2}{c}{\textbf{Average pooling}}\\ 
 &  Accuracy &  F1 & Accuracy & F1 & Accuracy  & F1 \\
\hline
Unsupervised      & 0.298    & 0.298 & 0.296 & 0.296 & 0.296 & 0.296  \\ \hline
Logistic regression           & 0.336    & 0.335   & 0.321 & 0.321 & 0.326 & 0.326    \\ \hline
CatBoost     & 0.318    & 0.318  & 0.318 & 0.318 & 0.329 & 0.330 \\
\hline
\end{tabular}
\label{tab:bert_mcqa_layers}

\begin{center}
\selectlanguage{english}
\caption{BERT embeddings results on NSP dataset (1st and 12th layer).}
\begin{tabular}{ c  c  c c c  c  c  c c c}

\hline
   \multirow{2}{*}{Method} & \multicolumn{2}{c}{\textbf{1st layer}} & \multicolumn{2}{c}{\textbf{12th layer}} &  \multicolumn{4}{c}{\textbf{Average pooling}}\\ 
 &  Accuracy &  F1 &   Accuracy  & F1 & Accuracy  & F1 \\
\hline
Unsupervised      & 0.429    & 0.429   & 0.508 & 0.508 & 0.484 & 0.484  \\ \hline
Logistic regression           & 0.244    & 0.244   & 0.248 & 0.248 & 0.247 & 0.246  \\ \hline
CatBoost     & 0.414    & 0.414   & 0.497 & 0.497 & 0.514 & 0.514 \\
\hline
\end{tabular}
\label{tab:bert_nsp_layers}
\end{center}

\begin{center}
\selectlanguage{english}
\caption{BERT embeddings results on PI dataset (1st and 12th layer).}
\begin{tabular}{ c  c  c  c  c  c  c  c }

\hline
   \multirow{2}{*}{Method} & \multicolumn{2}{c}{\textbf{1st layer}} & \multicolumn{2}{c}{\textbf{12th layer}} &  \multicolumn{2}{c}{\textbf{Average pooling}}\\ 
 & Accuracy  & F1 &   Accuracy  & F1 &  Accuracy  & F1 \\
\hline
Unsupervised      & 0.795    & 0.853 & 0.781 & 0.850 & 0.789 & 0.853  \\ \hline
Logistic regression           & 0.670  & 0.615 & 0.704 & 0.664 & 0.703 & 0.662   \\ \hline
CatBoost     & 0.762    & 0.708 & 0.758 & 0.706 & 0.764 & 0.718 \\
\hline
\end{tabular}
\label{tab:bert_para_layers}
\end{center}

\begin{center}
\selectlanguage{english}
\caption{Final results.}
\begin{tabular}{m{2cm} c c c c c c}

\hline
   \ \ \ \multirow{2}{*}{Model} & \multicolumn{2}{c}{\textbf{MCQA}} & \multicolumn{2}{c}{\textbf{NSP}} &  \multicolumn{2}{c}{\textbf{PI}}\\ 
 & \ Accuracy \ & \ \ F1 \  \  &  \ \  Accuracy \ \  & F1 & \ \  Accuracy \ \  & F1 \\
\hline
FastText     & 0.318    & 0.317 & 0.496 & 0.496 & 0.762 & 0.806  \\
ELMo     & 0.318    & 0.318 & 0.691 & 0.691 & 0.807 & 0.867  \\
BERT   & 0.346    & 0.346 & 0.514 & 0.514 & 0.801 & 0.857  \\
\hline
\end{tabular}
\label{tab:final}
\end{center}

\end{table}

The results of different methods based on BERT embeddings are shown in Tables \ref{tab:bert_mcqa} -- \ref{tab:bert_para_layers}. All the results were obtained using BERT--Base Multilingual Cased model\footnote{\url{https://github.com/google-research/bert/blob/master/multilingual.md}}. There were considered different BERT model layers and combinations of layers. For each method and task, there is presented the best result achieved with layer indication. From Table \ref{tab:bert_mcqa} we can see that using BERT embeddings we can significantly improve results in MCQA task. The CatBoost method based on BERT model noticeably outperforms FastText and ELMo models within this task. According to Table \ref{tab:bert_nsp}, we can claim that BERT model could not get better results compared to ELMo. Even the performance of logistic regression remained the same. However, as we can notice from Table \ref{tab:bert_para}, the performance of BERT model regarding paraphrase identification task is comparable to the ELMo results. In both cases, the best score was achieved by the unsupervised method and the worst by logistic regression. We suppose that in many cases BERT could not outperform ELMo model because it was not trained for these tasks and the best way for BERT is to fine--tune it by ourselves.


\section{Conclusion} 

We tested three sentence embedding models:  (a) FastText averaged over words in the sentence, b) three pre--trained on various sources ELMo models, c) BERT model in three NLU tasks for the Russian language. These tasks are a) multiple choice question answering, b) next sentence prediction, c) paraphrase identification. For the first two tasks, we presented our own new datasets. These datasets are designed as multiple choice questions: given a question / a sentence one need to choose a correct option from four possible answers / continuations. 

Our experiments show that the MCQA dataset is much more complicated than the other two datasets. The quality of the results for this task is not as high as for two others. All models perform somewhat similar in next sentence prediction and paraphrase identification tasks. The paraphrase identification dataset is highly unbalanced, and the positive examples are in the minority, which may affect the quality if the results. 

Overall, we can claim that we started to evaluate the popular sentence embeddings frameworks in GLUE--like fashion for the Russian language. So far we can state that (1) the word--level embeddings are outperformed by the sentence--level embeddings, (2) the pre--trained models available online with no doubts can attempt some of the NLU tasks with little or almost no fine tuning. The directions of the future work may include probing of embedding models for Russian rich morphology and free word order. The code of all experiments is available on GitHub\footnote{\url{https://github.com/foksly/aist-sentence-embeddings}}.

\section*{Acknowledgements}
This project was supported by the framework of the HSE University Basic Research Program and Russian Academic  Excellence Project ``5--100''. 
\bibliographystyle{splncs04}
\bibliography{papers}

\end{document}